\documentclass{article}
\usepackage{ijcai24}

\usepackage{times}
\usepackage{soul}
\usepackage{url}
\usepackage[hidelinks]{hyperref}
\usepackage[utf8]{inputenc}
\usepackage[small]{caption}
\usepackage{graphicx}
\usepackage{amsmath}
\usepackage{amsthm}
\usepackage{booktabs}
\usepackage[switch]{lineno}

\usepackage{subcaption}
\usepackage[ruled]{algorithm2e}
\SetKwInput{KwInput}{Input}                
\SetKwInput{KwOutput}{Output}              
\usepackage[table,xcdraw]{xcolor}
\definecolor{ylp_color1}{RGB}{255,193,193}
\definecolor{ylp_color2}{RGB}{255,228,225}
\usepackage{tcolorbox}
\newtcbox{\mybox}[1][red]{on line, colback = ylp_color2, colframe = ylp_color1,  arc=1mm, auto outer arc, boxrule=0.5pt,}
\usepackage{multirow}
\usepackage{booktabs}
\usepackage{wrapfig}
\usepackage{amsmath}
\usepackage{amssymb}
\usepackage{amsthm}
\usepackage{amsfonts}
\usepackage{bm}
\usepackage{eucal}
\usepackage{arydshln}

\newcommand{\ijcai}[1]{\color{black}{#1}}

\newcommand{\methodname}{{\tt{FedSSA}}}
\newtheorem{theorem}{Theorem}

\newtheorem{lemma}{Lemma}

\newtheorem{assumption}{Assumption}

\usepackage{pifont}
\usepackage{setspace}
\usepackage{url}

\title{FedSSA: Semantic Similarity-based Aggregation for \\ Efficient Model-Heterogeneous Personalized Federated Learning}

\author{
Liping Yi$^{1,2}$,
Han Yu$^2$,
Zhuan Shi$^{3}$,
Gang Wang$^{1,*}$, 
Xiaoguang Liu$^{1}$,  
Lizhen Cui$^{4,*}$, 
Xiaoxiao Li$^5$
\affiliations
$^1$College of Computer Science, TMCC, SysNet, DISSec, GTIISC, Nankai University, Tianjin, China\\
$^2$College of Computing and Data Science, Nanyang Technological University, Singapore\\
$^3$Artificial Intelligence Laboratory, École Polytechnique Fédérale de Lausanne (EPFL), Switzerland\\
$^4$School of Software, Shandong University, Jinan, China\\
$^5$Department of Electrical and Computer Engineering, The University of British Columbia, Vancouver, BC, Canada
\emails
\{yiliping, wgzwp, liuxg\}@nbjl.nankai.edu.cn,
han.yu@ntu.edu.sg,
zhuan.shi@epfl.ch,
clz@sdu.edu.cn, \\
xiaoxiao.li@ece.ubc.ca
}

\begin{document}

\maketitle

\begin{abstract}
Federated learning (FL) is a privacy-preserving collaboratively machine learning paradigm. Traditional FL requires all data owners (a.k.a. FL clients) to train the same local model. This design is not well-suited for scenarios involving data and/or system heterogeneity. Model-Heterogeneous Personalized FL (MHPFL) has emerged to address this challenge. Existing MHPFL approaches often rely on a public dataset with the same nature as the learning task, or incur high computation and communication costs. To address these limitations, we propose the \underline{Fed}erated \underline{S}emantic \underline{S}imilarity \underline{A}ggregation (\methodname{}) approach {\ijcai{for supervised classification tasks}}, which splits each client's model into a heterogeneous (structure-different) feature extractor and a homogeneous (structure-same) classification header. It performs local-to-global knowledge transfer via semantic similarity-based header parameter aggregation. 
In addition, global-to-local knowledge transfer is achieved via an adaptive parameter stabilization strategy which fuses the seen-class parameters of historical local headers with that of the latest global header for each client. \methodname{} does not rely on public datasets, while only requiring partial header parameter transmission to save costs. Theoretical analysis proves the convergence of \methodname{}. Extensive experiments present that \methodname{} achieves up to $3.62\%$ higher accuracy, $15.54$ times higher communication efficiency, and $15.52$ times higher computational efficiency compared to 7 state-of-the-art MHPFL baselines.
\end{abstract}

\section{Introduction}
As societies become increasingly aware of the importance of data privacy protection \cite{SU-Net}, centrally collecting large-scale data for model training has become less viable, especially under privacy regulations. To enable decentralized data owners to collaboratively train effective machine learning models while protecting privacy, federated learning (FL) \cite{FedAvg} has been proposed. In a typical FL system, a central FL server coordinates multiple data owners (FL clients) for model training. In each communication round, the server broadcasts its current global model to the clients. Each client then treats the received global model as its local model, and trains it on local private data. The updated local model is uploaded to the server. The server weighted averages (aggregates) the received local models to obtain an updated global model. The above steps are repeated until the global model converges. In FL, only model parameters are communicated between the server and clients. Potentially sensitive local data are not exposed, thereby protecting privacy \cite{FedRRA,FFEDCL,Yu-Federated-Graph}.


\begin{figure}[t]
\centering
\includegraphics[width=0.7\linewidth]{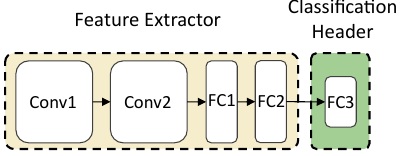}
\caption{{\ijcai{Feature extractor and classification header.}}}
\label{fig:Demo_extractor_header}
\vspace{-1em}
\end{figure}

This model homogeneous mode of FL requires all clients to train models with the same structures. It is not well-suited for scenarios involving data heterogeneity \cite{Non-IID} (i.e., clients with local data following non-independent and identical distributions (non-IID)) and system heterogeneity \cite{PruneFL,FedPE,Yu-radionetworks,yu-AnomalyDetection,D2D-LSTM,D2D-LSTM2,QM-RGNN} (i.e., clients with diverse communication and computation capabilities). Therefore, training a personalized FL model adaptively based on each client's local data distribution and system capability helps improve performances \cite{PFL-yu,pFedLHNs,pFedKT}. Furthermore, enterprise FL clients are often reluctant to share their model structures with others due to intellectual property concerns. Thus, the field of Model-Heterogeneous Personalized Federated Learning (MHPFL) \cite{FedGH,pFedES,FedLoRA,pFedMoE}, which allows each client to train a heterogeneous personalized local model with a structure tailored to actual needs, has emerged.

Existing MHPFL methods are mainly designed based on knowledge distillation, mutual learning, and model mixup. Knowledge distillation-based methods \cite{FedMD,FedDF,FCCL,DS-FL,CFD,FedHeNN,Cronus} often rely on the availability of a public dataset that is large enough and follows a similar distribution with local private data for fusing knowledge from heterogeneous local models. However, such a public dataset is difficult to obtain in practice. Other knowledge distillation-based methods \cite{FD,HFD1,HFD2,FedProto} without requiring public datasets often incur high computational overhead on clients and face the risk of privacy leakage. Mutual learning-based methods \cite{FML,FedKD} allow each client to train a large heterogeneous model and a small homogeneous model in a mutual learning manner. Only the small homogeneous models are uploaded to the server for aggregation. Training two models simultaneously leads to high computational overhead for clients. Model mixup-based methods \cite{FedRep,LG-FedAvg,FedClassAvg} divide each client's local model into a heterogeneous feature extractor and a homogeneous classifier. The server aggregates the homogeneous local classifiers to update the global classifier and then broadcasts it to clients. 
Each client replaces its local classifier with the updated global classifier or constructs a regularization term based on the distillation loss between local and global classifiers. 
They often ignore the semantic information of the classifier parameters, resulting in limited accuracy improvements.

To enable the privacy-preserving, communication- and computation-efficient training of high-performance models through MHPFL,
we propose the \underline{Fed}erated \underline{S}emantic \underline{S}imilarity \underline{A}ggregation (\methodname{}) approach
for supervised classification tasks, which have been widely applied in fields like cancer detection for medical diagnosis and object recognition in autonomous driving \cite{1w-survey}. {\ijcai{As depicted in Figure~\ref{fig:Demo_extractor_header},}}
it splits each client's local model into two parts: 1) a heterogeneous feature extractor, and 2) a homogeneous classification header. Local-to-global knowledge transfer is achieved by semantic similarity-based classification header parameter aggregation. 
Since the parameters of classification headers from different clients corresponding to the same class are semantically similar, each client only needs to upload classification header parameters corresponding to locally seen classes. The FL server aggregates parameters from different classification headers by class to update the global header. Global-to-local knowledge transfer is achieved through parameter stabilization. To ensure stable convergence, we devise an adaptive parameter stabilization strategy that fuses local historical header parameters and the latest global header parameters for locally seen classes to update the local header of each FL client. 

Owing to the above design, \methodname{} has the following advantages: 1) aggregating semantically similar head parameters corresponding to the same class stabilizes the decision boundaries and boosts performance; 2) the adaptive parameter stabilization strategy alleviates parameter shaking in the initial training rounds, thereby speeding up convergence; 3) the server and clients only need to transmit classification header parameters corresponding to seen classes, which is communication-efficient compared to transmitting the entire model; 4) low computational overhead for FL clients as they only fuse partial parameters from the global and local headers before local training; and 5) since only partial parameters are transmitted between the FL server and FL clients, no new privacy risk is introduced. 
In short, compared with existing methods, \methodname{} improves model performance, and communication and computation efficiency simultaneously.

Theoretical analysis proves the convergence of \methodname{}. Extensive experiments on two real-world datasets against seven compared with state-of-the-art MHPFL baselines demonstrate that \methodname{} achieves up to $3.62\%$ higher accuracy, $15.54$ times higher communication efficiency, and $15.52$ times higher computation efficiency.



\section{Related Work} \label{sec:related}
Our work is closely related to MHPFL with \textit{complete} model heterogeneity. These methods support flexible model heterogeneity, and are divided into the following three categories.

\textbf{Knowledge Distillation-based MHPFL}. Most knowledge distillation-based MHPFL methods rely on a public dataset (e.g., {\tt{FedMD}} \cite{FedMD}, {\tt{FedDF}} \cite{FedDF}, {\tt{FCCL}} \cite{FCCL}, {\tt{DS-FL}} \cite{DS-FL}, {\tt{CFD}} \cite{CFD}, {\tt{FedHeNN}} \cite{FedHeNN}, {\tt{Cronus}} \cite{Cronus}, {\tt{FSFL}} \cite{FSFL}, {\tt{FedAUX}} \cite{FEDAUX}, {\tt{FedKT}} \cite{FedKT}, {\tt{Fed-ET}} \cite{Fed-ET}, {\tt{FedKEMF}} \cite{FedKEMF}, {\tt{FedGEMS}} \cite{FedGEMS},  {\tt{KT-pFL}} \cite{KT-pFL}) to fuse information of local heterogeneous models by knowledge distillation. However, the public dataset is not always available. To avoid relying public datasets, {\tt{FedGen}} \cite{FedGen} and {\tt{FedZKT}} \cite{FedZKT} train a generator for generating local representations or public shared datasets, which is time-consuming and computation-intensive. In {\tt{FD}} \cite{FD}, {\tt{HFD}} \cite{HFD1,HFD2}, {\tt{FedProto}} \cite{FedProto}, {\tt{FedGKT}} \cite{FedGKT}, each client uploads the (class-average) logits or representations of local data to the server, the aggregated global logits or representations for each class are sent to clients for local knowledge distillation which incurs high computational costs for clients. Besides, uploading the logits/representations might compromise privacy. 

\textbf{Mutual Learning-based MHPFL}. {\tt{FML}} \cite{FML} and {\tt{FedKD}} \cite{FedKD} enable each client to train a large heterogeneous model and a small homogeneous model in a mutual learning manner. The large model is always trained locally and the small model is uploaded to the server for aggregation. Although they implement information interaction through the small homogeneous models, training the homogeneous model increases local computational costs for clients, and transmitting the homogeneous models incurs high communication costs. 

\textbf{Model Mixup-based MHPFL}. These methods split each local model into a feature extractor and a classifier. In {\tt{FedRep}} \cite{FedRep}, {\tt{FedPer}} \cite{FedPer}, {\tt{FedMatch}} \cite{FedMatch}, {\tt{FedBABU}} \cite{FedBABU} and {\tt{FedAlt/FedSim}} \cite{FedAlt/FedSim}, the feature extractor is homogeneous and used for aggregation by the FL server to enhance generalization. The classifier can be heterogeneous. Since the feature extractor has more parameters than the classifier, these methods can only support model heterogeneity to a low degree. In contrast, {\tt{LG-FedAvg}} \cite{LG-FedAvg}, {\tt{FedClassAvg}} \cite{FedClassAvg} and {\tt{CHFL}} \cite{CHFL} use heterogeneous feature extractors and homogeneous classifiers (i.e., executing the same classification task). The local classifiers are uploaded to the FL server for aggregation to generate the global classifier. To acquire global knowledge, {\tt{LG-FedAvg}} directly replaces each client's local classifier with the global classifier. Each client in {\tt{CHFL}} or {\tt{FedClassAvg}} calculates the regularization term or distillation loss between the local and global classifiers. Although these methods support a higher degree of model heterogeneity, they ignore the semantic similarity of classifier parameters belonging to the same class, thus achieving limited performance improvement. Our \methodname{} sets out to address the aforementioned limitations.


 \section{Preliminaries}
\subsection{Notations and Objective of Typical FL} 
A typical FL system involves a central FL server and $N$ decentralized FL clients. In each round, the server selects a $C$ fraction of clients at random. The selected client set is denoted as $\mathcal{K}$,  $|\mathcal{K}|=C \cdot N=K$. The server then broadcasts the global model $f(\omega)$ ($f(\cdot)$ and $\omega$ denote the model structure and the model parameters) to the selected clients. A client $k$ trains the received $f(\omega)$ on its local dataset $D_k$ to produce a local model $f(\omega_k)$ through gradient descent $\omega_k \gets \omega-\eta \nabla \ell(f(\boldsymbol{x}_i; \omega), y_i)$. $\ell(f(\boldsymbol{x}_i; \omega), y_i)$ is the loss of the global model $f(\omega)$ on the sample $(\boldsymbol{x}_i,y_i) \in D_k$. $D_k \sim P_k$ indicates that $D_k$ obeys distribution $P_k$ (i.e., local data from different clients are non-IID). Then, client $k$ uploads its local model parameters $\omega_k$ to the server. The server aggregates the received local models to update the global model, $\omega=\sum_{k=0}^{K-1} \frac{n_k}{n} \omega_k$. That is, the objective of typical FL is to minimize the average loss of the global model $f(\omega)$ on data from all clients:
\begin{equation}
\min _{\omega \in \mathbb{R}^d} \sum_{k=0}^{K-1} \frac{n_k}{n} \mathcal{L}_k(D_k ; f(\omega)),
\end{equation}
where $\omega$ are $d$-dimensional real numbers. $n_k=|D_k|$ is the number of samples in client $k$'s local dataset. $n=\sum_{k=0}^{N-1} n_k$. $\mathcal{L}_k(D_k ; f(\omega))=\frac{1}{|D_k|} \sum_{(\boldsymbol{x}_i, y_i) \in D_k} \ell(f(\boldsymbol{x}_i ; \omega), y_i)$ is the loss of the global model $f(\omega)$ on $D_k$.


\subsection{Problem Definition}
The problem we aim to solve in this work belongs to the category of MHPFL for \textit{supervised classification} tasks. Each FL client $k$ owns local models $f_k(\omega_k)$ with a model structure $f_k(\cdot)$ and parameters $\omega_k$. They can be heterogeneous for different clients.
We assume all clients execute the same classification task and
each client's local model $f_k(\omega_k)$ consists of a heterogeneous feature extractor $\mathcal{F}_k(\varphi_k)$ and a homogeneous classification header $\mathcal{H}(\theta_k)$, i.e., $f_k(\omega_k)=\mathcal{F}_k(\varphi_k) \circ \mathcal{H}(\theta_k)$ with $\circ$ denoting model splicing. 
The feature extractor takes inputs as $\mathcal{F}_k(D_k;\varphi_k)$, where $\mathcal{F}_k(\cdot), \varphi_k$ denote the structure and parameters of the feature extractor. It maps data $D_k$ from the input space to the feature space. 
The classification header takes input as $\mathcal{H}(\mathcal{F}_k(D_k;\varphi_k);\theta_k)$, where $\mathcal{H}(\cdot), \theta_k$ denote the structure and parameters of the classification header. It involves the last two linear layers of the model which maps the features $\mathcal{F}_k(D_k;\varphi_k)$ extracted by $\mathcal{F}_k(\cdot)$ to the output space. 

Our objective is to minimize the sum of losses of all clients' heterogeneous local models:
\begin{equation}
\small
\min _{\omega_0, \ldots, \omega_{K-1}} \sum_{k=0}^{K-1} \mathcal{L}_k(D_k ; f_k(\omega_k))=\sum_{k=0}^{K-1} \mathcal{L}_k(D_k ; \mathcal{F}_k(\varphi_k) \circ \mathcal{H}(\theta_k)),
\end{equation}
where local models $\omega_0, \ldots, \omega_{K-1} \in \mathbb{R}^{d_0, \ldots, d_{K-1}}$.

\section{The Proposed \methodname{} Approach}
\begin{figure}[b]
\centering
\includegraphics[width=0.9\linewidth]{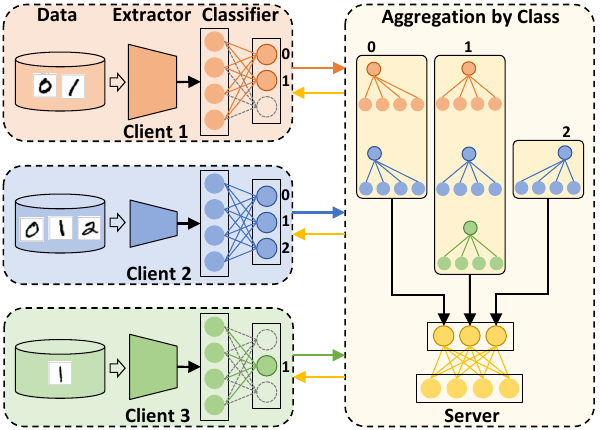}
\caption{The \methodname{} framework.}
\label{fig:FedSSA}
\end{figure}

\methodname{} consists of two modules: 1) semantic similarity-based classification header parameter aggregation for local-to-global knowledge transfer, and 2) adaptive parameter stabilization for global-to-local knowledge transfer.

\subsection{Semantic Similarity-Based Aggregation}

As shown in Figure~\ref{fig:FedSSA}, the last fully connected layer of the classification header outputs all-class logits which are mapped to the prediction probability for each class by a soft-max layer.
That is, each neuron at the classification header's last layer corresponds to one class and its connections with all neurons at the previous layer are the parameters belonging to corresponding classes.
Assuming that all clients execute the same $S$-classification task, then client $k$'s local classification header $\theta_k$ can be subdivided by class as $\{\theta_k^0,\dots,\theta_k^s,\dots,\theta_k^{S-1}\}$. Since local data from different clients are non-IID, different clients may hold partial classes of data. We denote the set of client $k$'s seen classes as $S_k$.

Since each client trains its local model on locally seen classes, the local classification header seen-class parameters are well-trained. Therefore, client $k$ only uploads its classification header seen-class parameters $\{\theta_k^s\}, s \in S_k$ to the FL server for aggregation. As illustrated in Figure~\ref{fig:FedSSA}, client $1$ only uploads its classification header parameters corresponding to seen classes `$0,1$', client $2$ only uploads its classification header parameters for classes `$0,1,2$', and client $3$ only uploads its classification header parameters for class `$1$'.

Since classification header parameters for the same class from different clients are \textit{semantically similar}, we design the server to aggregate classification header parameters based on classes (i.e., the corresponding neuron positions at the output layer). We denote the set of clients holding class $s$ of data as $\mathcal{K}_s$, and define the aggregation rule as: 
\begin{equation}
\theta^s=\frac{1}{|\mathcal{K}_s|} \sum_{k \in \mathcal{K}_s} \theta_k^s.
\end{equation}
After aggregation, the server splices the aggregated header parameters for all classes together to form an updated global header $\theta=\{\theta^0,\dots,\theta^s,\dots,\theta^{S-1}\}$. Then, it sends the partial parameters of the global header to each client corresponding to its respective locally seen classes. This step transfers local seen-class knowledge from clients to the server. It enhances the generalization to seen classes for each client after receiving the aggregated seen-class header parameters, and produces a more stable seen-class classification boundary.

\begin{figure}[b]
\centering
\includegraphics[width=0.55\linewidth]{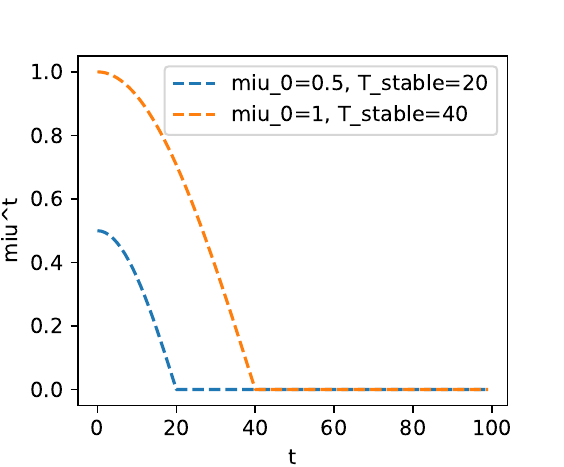}
\caption{Decay functions for $\mu^t$. $\cos (\cdot)\in[0,\pi/2]$ is a smooth decay function. It leads to a stable decrease of $\mu$.}
\label{fig:miu_function}
\end{figure}

\subsection{Adaptive Parameter Stabilization}
Inspired by historical learning \cite{historical-learning} {\ijcai{which tends to re-use historical knowledge from old models to avoid forgetting}}, during global-to-local knowledge transfer, we exploit historical local headers to stabilize and speed up model convergence. Specifically, in the $t$-th training round, client $k$ fuses the seen-class parameters from its historical local header $\theta_k^{t-1,s}$ ($s\in S_k$) and the latest global header $\theta^{t-1,s}$ ($s\in S_k$) to produce a new local header $\hat{\theta}_k^{t,s}$. The unseen-class header parameters are still the historical local header $\theta_k^{t-1,s^{\prime}}$ ($s^{\prime} \in (S \backslash S_k)$). The fusion of the seen-class parameters from the two headers is achieved by the proposed adaptive parameter stabilization strategy.

In the initial rounds of training, local model parameters are updated rapidly and local model parameters from different clients trained on non-IID data are biased. This leads to instability in the global header. To enhance the reliability of global-to-local knowledge transfer, we devise an adaptive parameter stabilization strategy: 
\begin{equation}\label{eq:para-stable}
\begin{aligned}
& \hat{\theta}_k^{t, s}=\theta^{t-1, s}+\mu^t \cdot \theta_k^{t-1, s}, s \in S_k, \\
& \mu^t= \begin{cases}\mu_0 \cdot \cos (\frac{t}{2 T_{stable }} \pi), & t \leq T_{stable}, \\
0, & t>T_{stable}.\end{cases} 
\end{aligned}
\end{equation}
The coefficient $\mu^t \in [0,1] $ of the historical local header is determined by the above piecewise function. In round $t \leq T_{stable}$ (a fixed hyperparameter for all clients in FL), $\mu^t$ is determined by a decay function (e.g., Figure~\ref{fig:miu_function}) with initial value $\mu_0 \in (0,1]$. $\mu^t$ decreases as round $t$ rises. When $t>T_{stable}$, the reliability of the global header has been enhanced. Thus, we set $\mu^t=0$ to directly replace historical local seen-class parameters with the latest global seen-class header parameters. This prevents stateless local header parameters from slowing down convergence.

The updated local header $\hat{\theta}_k^t=\{\hat{\theta}_k^{t, s}, \theta_k^{t-1, s^{\prime}}\}(s \in S_k, s^{\prime} \in (S \backslash S_k))$ and the feature extractor $\varphi_k^{t-1}$ from the previous round are spliced together to form a new local model $\hat{\omega}_k^t$:
\begin{equation}
    \hat{\omega}_k^t = \varphi_k^{t-1} \circ \hat{\theta}_k^t.
\end{equation}
Then, $\hat{\omega}_k^t$ is trained on $D_k$ to obtain $\omega_k^t$ via gradient descent:
\begin{equation}
\omega_k^t \gets \hat{\omega}_k^t-\eta \nabla(f_k(\boldsymbol{x}_i ; \hat{\omega}_k^t), y_i),(\boldsymbol{x}_i, y_i) \in D_k.
\end{equation}

These three key steps of \textit{uploading seen-class header parameters by class}, \textit{aggregation by class}, and \textit{parameter fusion by class} enhance the personalization and stabilize the classification boundary for each heterogeneous local model. They are repeated until local models from all clients converge. Each client's final local model is used for inference. \methodname{} is detailed in Algorithm~\ref{alg:FedSSA} (Appendix \ref{app:alg}).

\subsection{Discussion}
Here, we discuss the privacy, communication costs, and computational overheads of \methodname{}.

\textbf{Privacy.} 
When a client uploads its seen-class header parameters, it can maintain the seen-class header parameters and replace the unseen-class header parameters with $0$, while still uploading the entire header to the server. The server aggregates {\ijcai{non-zero}} local header parameters according to the corresponding classes (ordinates), and then broadcasts the updated global header to the clients. Since only the header parameters are transmitted, local data privacy can be preserved.

\textbf{Communication Cost.}
As stated above, only the parameters of headers are communicated between clients and the server in each round of FL. Therefore, \methodname{} incurs lower communication costs than transmitting the complete models as in the cases of {\tt{FedAvg}} based FL approaches.

\textbf{Computational overhead.} For clients, the additional computation compared with {\tt{FedAvg}} is incurred by header parameter fusion in the adaptive parameter stabilization strategy, which are simple linear operations. It is significantly lower than the computational overhead incurred by training local models. For the server, aggregating the header parameters by class linearly incurs low computation costs.

\section{Analysis}
To analyze the convergence of \methodname{}, we first introduce some additional notations. Let $e \in \{0, 1, ..., E\}$ denote a local iteration. $(tE+e)$ denotes the $e$-th local iteration in the FL training round $(t+1)$. At $(tE+0)$ (i.e., the beginning of the round $(t+1)$), clients fuse historical local headers (in round $t$) and the latest global header (in round $t$) to update local headers. $(tE+1)$ is the first iteration in round $(t+1)$. $(tE+E)$ is the last iteration in round $(t+1)$. 

\begin{assumption}\label{assump:Lipschitz}
\textbf{Lipschitz Smoothness}. Client $k$'s local model gradients are $L1$--Lipschitz smooth, i.e.,
\begin{equation}\label{eq:7}
\footnotesize
\begin{gathered}
\|\nabla \mathcal{L}_k^{t_1}(\omega_k^{t_1} ; \boldsymbol{x}, y)-\nabla \mathcal{L}_k^{t_2}(\omega_k^{t_2} ; \boldsymbol{x}, y)\| \leqslant L_1\|\omega_k^{t_1}-\omega_k^{t_2}\|, \\
\forall t_1, t_2>0, k \in\{0,1, \ldots, N-1\},(\boldsymbol{x}, y) \in D_k.
\end{gathered}
\end{equation}
From Eq. \eqref{eq:7}, we further derive:
\begin{equation}
\footnotesize
\mathcal{L}_k^{t_1}-\mathcal{L}_k^{t_2} \leqslant\langle\nabla \mathcal{L}_k^{t_2},(\omega_k^{t_1}-\omega_k^{t_2})\rangle+\frac{L_1}{2}\|\omega_k^{t_1}-\omega_k^{t_2}\|_2^2 .
\end{equation}

\end{assumption}

\begin{assumption} \label{assump:Unbiased}
\textbf{Unbiased Gradient and Bounded Variance}. The random gradient $g_k^t=\nabla \mathcal{L}_k^t(\omega_k^t; \mathcal{B}_k^t)$ ($\mathcal{B}$ is a batch of local data) of each client's local model is unbiased, i.e.,
\begin{equation}
\footnotesize
\mathbb{E}_{\mathcal{B}_k^t \subseteq D_k}[g_k^t]=\nabla \mathcal{L}_k^t(\omega_k^t),
\end{equation}
and the variance of random gradient $g_k^t$ is bounded by:
\begin{equation}
\footnotesize
\mathbb{E}_{\mathcal{B}_k^t \subseteq D_k}[\|\nabla \mathcal{L}_k^t(\omega_k^t ; \mathcal{B}_k^t)-\nabla \mathcal{L}_k^t(\omega_k^t)\|_2^2] \leqslant \sigma^2 .
\end{equation}    

\end{assumption} 

\begin{assumption} \label{assump:header}
\textbf{Bounded Variance of Classification Headers}. The variance between the seen-class parameters $\theta_k^s (s\in S_k)$ of client $k$'s local header $\mathcal{H}(\theta_k)$ and the same seen-class parameters $\theta^s=\mathbb{E}_{k^{\prime} \in \mathcal{K}_s}[\theta_{k^{\prime}}^s] (s\in S_k)$ of the global header $\mathcal{H}(\theta)$ is bounded \cite{FedGH}:

\textit{Parameter bounded}: $\mathbb{E}_{s \in S_k}[\|\theta^s-\theta_k^s\|_2^2] \leq \varepsilon^2$,

\textit{Gradient bounded}: $\mathbb{E}_{s \in S_k}[\|\nabla \mathcal{L}(\theta^s)-\nabla \mathcal{L}(\theta_k^s)\|_2^2] \leq \delta_1^2$.
\end{assumption}

With the above assumptions, since \methodname{} does not change local model training, Lemma \ref{lemma:LocalTraining} from \cite{FedProto} still holds.

\begin{lemma} \label{lemma:LocalTraining}
Based on Assumptions~\ref{assump:Lipschitz} and \ref{assump:Unbiased}, during $\{0,1,...,E\}$ iterations of the $(t+1)$-th FL training round, the loss of an arbitrary client's local model is bounded by:
\begin{equation}
\footnotesize
\mathbb{E}[\mathcal{L}_{(t+1) E}] \leqslant \mathcal{L}_{t E+0}-(\eta-\frac{L_1 \eta^2}{2}) \sum_{e=0}^E\|\mathcal{L}_{t E+e}\|_2^2+\frac{L_1 E \eta^2}{2} \sigma^2 .
\end{equation}
\end{lemma}

\begin{lemma} \label{lemma:AfterAggregation}
Based on Assumption~\ref{assump:header}, the loss of an arbitrary client's local model after fusing the seen-class parameters of local and global headers by the adaptive parameter stabilization strategy is bounded by:
\begin{equation}
\footnotesize
\mathbb{E}[\mathcal{L}_{(t+1) E+0}] \leqslant \mathbb{E}[\mathcal{L}_{(t+1) E}]-\frac{\eta L_1 \delta^2}{2}.
\end{equation}
\end{lemma}
The detailed proof can be found in Appendix~\ref{sec:proof-Lemma2}.

Based on Lemma~\ref{lemma:LocalTraining} and Lemma \ref{lemma:AfterAggregation}, we can further derive the following theorems.

\begin{theorem} \label{theorem:one-round}
\textbf{One-round deviation}. The expectation of the loss of an arbitrary client's local model before the beginning of a round of local iteration satisfies
\begin{equation}
\footnotesize
\begin{aligned}
\mathbb{E}[\mathcal{L}_{(t+1) E+0}] \leq & \mathcal{L}_{t E+0}-(\eta-\frac{L_1 \eta^2}{2}) \sum_{e=0}^E\|\mathcal{L}_{t E+e}\|_2^2 \\
& \quad +\frac{\eta L_1(E \eta \sigma^2-\delta^2)}{2}.
\end{aligned}
\end{equation}
\end{theorem}

The proof can be found in Appendix~\ref{sec:proof-Theorem1}. Then we can derive the non-convex convergence rate of \methodname{} as follows:

\begin{theorem} \label{theorem:non-convex}
\textbf{Non-convex convergence rate of \methodname{}}. Based on the above assumptions and derivations, for an arbitrary client and any $\epsilon>0$, the following inequality holds:
\begin{equation}
\footnotesize
\begin{aligned}
\frac{1}{T} \sum_{t=0}^{T-1} \sum_{e=0}^E \mathbb{E}[\|\mathcal{L}_{t E+e}\|_2^2] & \leqslant  \frac{2(\mathcal{L}_{t=0}-\mathcal{L}^*)}{T \eta(2-L_1 \eta)}+\frac{L_1(E \eta \sigma^2-\delta^2)}{2-L_1 \eta} 
\leqslant \epsilon, \\
\text { s.t. } \eta & < \frac{2 \epsilon+L_1 \delta^2}{L_1(\epsilon+E \sigma^2)}.
\end{aligned}
\end{equation}
\end{theorem}
Therefore, under \methodname{}, an arbitrary client's local model converges at a non-convex convergence rate of $\epsilon \sim \mathcal{O}(\frac{1}{T})$. The detailed proof can be found in Appendix~\ref{sec:proof-Theorem2}.

\section{Experimental Evaluation}
In this section, we compare \methodname{} \footnote{\url{https://github.com/LipingYi/FedSSA}} against $7$ state-of-the-art MHPFL approaches on two real-world datasets under various experiment conditions. 
and baselines with Pytorch on four NVIDIA GeForce RTX 3090 GPUs with 24G memory. 

\subsection{Experiment Setup}
\textbf{Datasets and Models}. We evaluate \methodname{} and baselines on two image classification datasets: CIFAR-10 and CIFAR-100 \footnote{\url{https://www.cs.toronto.edu/\%7Ekriz/cifar.html}} \cite{cifar}. They are manually divided into non-IID datasets following the method specified in \cite{pFedHN}. For CIFAR-10, we assign only data from 2 out of the 10 classes to each client (non-IID: 2/10). For CIFAR-100, we assign only data from 10 out of the 100 classes to each client (non-IID: 10/100). Then, each client's local data are further divided into the training set, the evaluation set, and the testing set following the ratio of 8:1:1. This way, the testing set is stored locally by each client, which follows the same distribution as the local training set. As shown in Table~\ref{tab:model-structures} (Appendix \ref{app:extra-exp}), each client trains a CNN model with output layer dimensions (i.e., the last fully-connected layer) of $10$ or $100$ on CIFAR-10 and CIFAR-100 datasets, respectively. The dimensions of the representation layer (i.e., the second last fully-connected layer) are set to be $500$. Therefore, each classification header consists of $10 (100)\times 500$ parameters.

\textbf{Baselines}. We compare \methodname{} with 7 baseline methods, which are the state-of-the-art methods in the three MHPFL categories elaborated in Sec.~\ref{sec:related}.

\begin{itemize}
    \item {\tt{Standalone}}, each client trains its local model independently, which serves as a lower performance bound; 
    \item Knowledge distillation without public data: {\tt{FD}} \cite{FD}, {\tt{FedProto}} \cite{FedProto}; 
    \item Mutual learning: {\tt{FML}} \cite{FML} and {\tt{FedKD}} \cite{FedKD};
    \item Model mixup: {\tt{LG-FedAvg}} \cite{LG-FedAvg} and {\tt{FedClassAvg}} \cite{FedClassAvg}.
\end{itemize}

\textbf{Evaluation Metrics}. \textbf{1) Accuracy}: we measure the \textit{individual test accuracy} ($\%$) of each client's local model and calculate the \textit{average test accuracy} of all clients' local models. \textbf{2) Communication Cost}: We trace the number of transmitted parameters when the average model accuracy reaches the target accuracy. \textbf{3) Computation Cost}: We track the consumed computation FLOPs when the average model accuracy reaches the target accuracy.

\textbf{Training Strategy}. We tune the optimal FL settings for all methods via grid search. The epochs of local training $E \in \{1, 10\}$ and the batch size of local training $B \in \{64, 128, 256, 512\}$. The optimizer for local training is SGD with learning rate $\eta=0.01$. We also tune special hyperparameters for the baselines and report the optimal results. We also adjust the hyperparameters $\mu_0$ and $T_{stable}$ to achieve the best-performance \methodname{}. To compare \methodname{} with the baselines fairly, we set the total number of communication rounds $T \in \{100, 500\}$ to ensure all algorithms converge.

\subsection{Comparisons Results}
We compare \methodname{} with the baselines under both model-homogeneous and model-heterogeneous scenarios 
with different total numbers of clients $N$ and client participation fraction $C$ to evaluate the robustness of \methodname{}. We set up three scenarios: $\{(N=10, C=100\%), (N=50, C=20\%), (N=100, C=10\%)\}$. For ease of comparison across the three settings, $N\times C$ is set to be the same (i.e., $10$ clients participate in each round of FL).

\subsubsection{Model-Homogeneous PFL}
In this scenario, all clients hold the largest model `CNN-1' as shown in Table~\ref{tab:model-structures}. Table~\ref{tab:exp-comparison-homo} shows that \methodname{} consistently achieves the highest mean test accuracy under all settings, outperforming the best baseline by $3.61\%$.

\begin{table}[t]
\centering
\caption{The mean test accuracy for \textit{model-homogeneous} scenarios. $N$ is the total number of clients. $C$ is the fraction of participating clients in each round. `-' denotes failure to converge.}
\label{tab:exp-comparison-homo}
\resizebox{1\linewidth}{!}{%
\begin{tabular}{|l|cc|cc|cc|}
\hline
            & \multicolumn{2}{c|}{N=10, C=100\%}                                                                   & \multicolumn{2}{c|}{N=50, C=20\%}                                                                    & \multicolumn{2}{c|}{N=100, C=10\%}                                                                   \\ \cline{2-7} 
Method      & \multicolumn{1}{c|}{CIFAR10}                               & CIFAR100                              & \multicolumn{1}{c|}{CIFAR10}                               & CIFAR100                              & \multicolumn{1}{c|}{CIFAR10}                               & CIFAR100                              \\ \hline
{\tt{Standalone}}  & \multicolumn{1}{c|}{96.35}                                  & 74.32                                  & \multicolumn{1}{c|}{\cellcolor[HTML]{D3D3D3}95.25}          & 62.38                                  & \multicolumn{1}{c|}{\cellcolor[HTML]{D3D3D3}92.58}          & \cellcolor[HTML]{D3D3D3}54.93          \\
{\tt{FML}}         & \multicolumn{1}{c|}{94.83}                                  & 70.02                                  & \multicolumn{1}{c|}{93.18}                                  & 57.56                                  & \multicolumn{1}{c|}{87.93}                                  & 46.20                                  \\
{\tt{FedKD}}       & \multicolumn{1}{c|}{94.77}                                  & 70.04                                  & \multicolumn{1}{c|}{92.93}                                  & 57.56                                  & \multicolumn{1}{c|}{90.23}                                  & 50.99                                  \\
{\tt{LG-FedAvg}}   & \multicolumn{1}{c|}{\cellcolor[HTML]{D3D3D3}96.47}          & 73.43                                  & \multicolumn{1}{c|}{94.20}                                  & 61.77                                  & \multicolumn{1}{c|}{90.25}                                  & 46.64                                  \\
{\tt{FedClassAvg}} & \multicolumn{1}{c|}{96.44}                                  & \cellcolor[HTML]{D3D3D3}74.33          & \multicolumn{1}{c|}{94.45}                                  & \cellcolor[HTML]{D3D3D3}63.10          & \multicolumn{1}{c|}{91.03}                                  & 49.31                                  \\
{\tt{FD}}          & \multicolumn{1}{c|}{96.30}                                  & -                                      & \multicolumn{1}{c|}{-}                                      & -                                      & \multicolumn{1}{c|}{-}                                      & -                                      \\
{\tt{FedProto}}   & \multicolumn{1}{c|}{95.83}                                  & 72.79                                  & \multicolumn{1}{c|}{95.10}                                  & 62.55                                  & \multicolumn{1}{c|}{91.19}                                  & 54.01                                  \\ \hline
\methodname{}      & \multicolumn{1}{c|}{\cellcolor[HTML]{9B9B9B}\textbf{98.57}} & \cellcolor[HTML]{9B9B9B}\textbf{74.81} & \multicolumn{1}{c|}{\cellcolor[HTML]{9B9B9B}\textbf{95.32}} & \cellcolor[HTML]{9B9B9B}\textbf{64.38} & \multicolumn{1}{c|}{\cellcolor[HTML]{9B9B9B}\textbf{92.67}} & \cellcolor[HTML]{9B9B9B}\textbf{58.54} \\ \hline
\end{tabular}%
}
\end{table}

\begin{table}[t]
\centering
\caption{The mean test accuracy for \textit{model-heterogeneous} scenarios. $N$ is the total number of clients. $C$ is the fraction of participating clients in each round. `-' denotes failure to converge.}
\label{tab:exp-comparison-hetero}
\resizebox{1\linewidth}{!}{%
\begin{tabular}{|l|cc|cc|cc|}
\hline
            & \multicolumn{2}{c|}{N=10, C=100\%}                                                                   & \multicolumn{2}{c|}{N=50, C=20\%}                                                                    & \multicolumn{2}{c|}{N=100, C=10\%}                                                                   \\ \cline{2-7} 
Method      & \multicolumn{1}{c|}{CIFAR10}                               & CIFAR100                              & \multicolumn{1}{c|}{CIFAR10}                               & CIFAR100                              & \multicolumn{1}{c|}{CIFAR10}                               & CIFAR100                              \\ \hline
{\tt{Standalone}}  & \multicolumn{1}{c|}{\cellcolor[HTML]{D3D3D3}96.53}          & 72.53                                  & \multicolumn{1}{c|}{95.14}                                  & \cellcolor[HTML]{D3D3D3}62.71          & \multicolumn{1}{c|}{91.97}                                  & 53.04                                  \\
{\tt{FML}}         & \multicolumn{1}{c|}{30.48}                                  & 16.84                                  & \multicolumn{1}{c|}{-}                                      & 21.96                                  & \multicolumn{1}{c|}{-}                                      & 15.21                                  \\
{\tt{FedKD}}       & \multicolumn{1}{c|}{80.20}                                  & 53.23                                  & \multicolumn{1}{c|}{77.37}                                  & 44.27                                  & \multicolumn{1}{c|}{73.21}                                  & 37.21                                  \\
{\tt{LG-FedAvg}}   & \multicolumn{1}{c|}{96.30}                                  & 72.20                                  & \multicolumn{1}{c|}{94.83}                                  & 60.95                                  & \multicolumn{1}{c|}{91.27}                                  & 45.83                                  \\
{\tt{FedClassAvg}} & \multicolumn{1}{c|}{94.41}                                  & 69.16                                  & \multicolumn{1}{c|}{85.23}                                  & 56.54                                  & \multicolumn{1}{c|}{82.61}                                  & 35.48                                  \\
{\tt{FD}}          & \multicolumn{1}{c|}{96.21}                                  & -                                      & \multicolumn{1}{c|}{-}                                      & -                                      & \multicolumn{1}{c|}{-}                                      & -                                      \\
{\tt{FedProto}}    & \multicolumn{1}{c|}{96.51}                                  & \cellcolor[HTML]{D3D3D3}72.59          & \multicolumn{1}{c|}{\cellcolor[HTML]{D3D3D3}95.48}          & 62.69                                  & \multicolumn{1}{c|}{\cellcolor[HTML]{D3D3D3}92.49}          & \cellcolor[HTML]{D3D3D3}53.67          \\ \hline
\methodname{}     & \multicolumn{1}{c|}{\cellcolor[HTML]{9B9B9B}\textbf{96.54}} & \cellcolor[HTML]{9B9B9B}\textbf{73.39} & \multicolumn{1}{c|}{\cellcolor[HTML]{9B9B9B}\textbf{95.83}} & \cellcolor[HTML]{9B9B9B}\textbf{64.20} & \multicolumn{1}{c|}{\cellcolor[HTML]{9B9B9B}\textbf{92.92}} & \cellcolor[HTML]{9B9B9B}\textbf{57.29} \\ \hline
\end{tabular}%
}
\end{table}

\subsubsection{Model-Heterogeneous PFL}
We evenly assign $5$ models {\ijcai{(model IDs: $1-5$)}} with different structures (Table~\ref{tab:model-structures}, Appendix \ref{app:extra-exp}) to clients. In practice, the id of the model assigned to client $k$ is determined by (client ID $k\% 5$). For {\tt{FML}} and {\tt{FedKD}}, we regard the $5$ structurally different models as large heterogeneous models and consider the smallest `CNN-5' model as the small homogeneous model.

\textbf{Mean Accuracy}. Table~\ref{tab:exp-comparison-hetero} shows that \methodname{} consistently achieves the highest mean test accuracy under all settings, outperforming the best baseline by $3.62\%$. Figure~\ref{fig:compare-hetero-converge} (Appendix \ref{app:extra-exp}) shows how the mean test accuracy of \methodname{} and the two top-performing baselines in each FL setting of Table~\ref{tab:exp-comparison-hetero} varies with communication round, it presents that \methodname{} converges to higher accuracies faster than others.


\begin{figure}[t]
\centering
\begin{minipage}[t]{0.25\textwidth}
\centering
\includegraphics[width=1.9in]{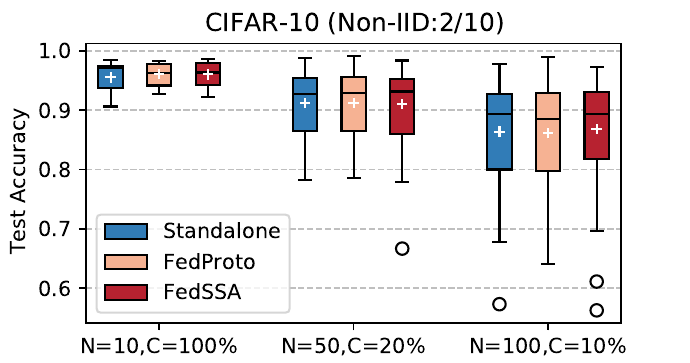}
\end{minipage}%
\begin{minipage}[t]{0.25\textwidth}
\centering
\includegraphics[width=1.9in]{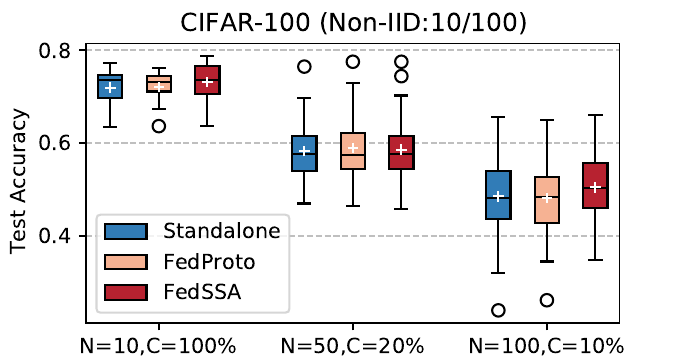}
\end{minipage}%
\caption{Accuracy distribution for individual clients.}
\label{fig:case-hetero-individual}
\vspace{-1em}
\end{figure}

\textbf{Individual Accuracy}. Figure~\ref{fig:case-hetero-individual} shows the accuracy distribution of individual models in \methodname{} and the two top-performing baselines in each FL setting shown in Table~\ref{tab:exp-comparison-hetero}.
In Figure~\ref{fig:case-hetero-individual}, `+' denotes the average accuracy of all clients under each algorithm. A small box length bounded by the upper quartile and the lower quartile indicates a more concentrated accuracy distribution across all clients with small variance. We observe that the three algorithms achieve similar mean and variance values in terms of accuracy when $N=\{10, 50\}$, while \methodname{} achieves significantly higher mean and lower variance when $N=100$. This verifies that \methodname{} is capable of producing the best personalized heterogeneous models in FL with a large number of potential clients and a low participation rate, which closely reflects practical FL scenarios.

\begin{figure}[t]
\centering
\begin{minipage}[t]{0.5\linewidth}
\centering
\includegraphics[width=1.7in]{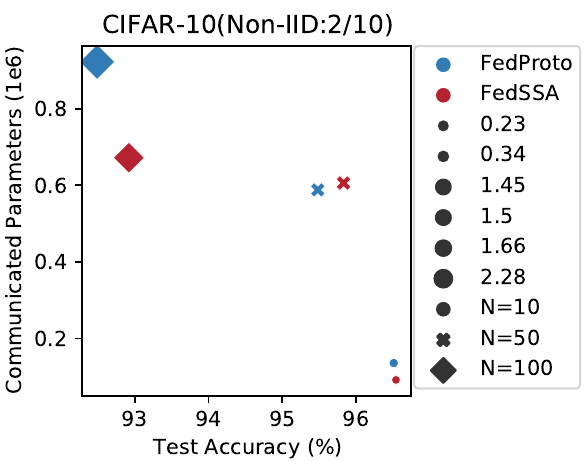}
\end{minipage}%
\begin{minipage}[t]{0.5\linewidth}
\centering
\includegraphics[width=1.7in]{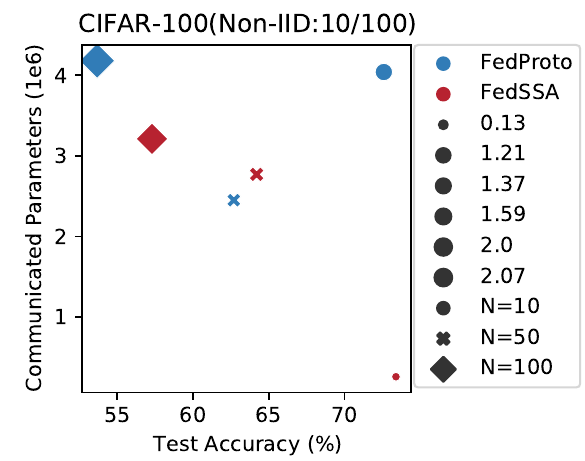}
\end{minipage}%
\vspace{-0.5em}
\caption{Trade-off among test accuracy, communication cost, and computational overhead. The sizes of each marker {\ijcai{(dots of varying sizes)}} reflect the corresponding computation FLOPs (1e12).}
\label{fig:compare-tradeoff}
\end{figure}



\textbf{Trade-off Among Accuracy, Communication and Computational costs}. Figure~\ref{fig:compare-tradeoff} reflects the trade-offs achieved by the comparison algorithms among accuracy, communication costs, and computational costs. The marker closer to the bottom-right-hand corner (higher accuracy and lower communication cost) of the figure and with a smaller size (lower computational cost) achieves the best trade-offs. Under all settings of (N, C) (markers with different types), \methodname{} achieves higher accuracies, and lower or similar communication and computational costs compared to the best-performing baseline, {\tt{FedProto}}, thus demonstrating its capability of achieving the most advantageous trade-off.
{\ijcai{Since we still upload the entire processed header to the server. One header has more parameters than the uploaded average representations of seen classes in {\tt{FedProto}}, \emph{i.e.,} \methodname{} has a higher per-round communication cost. Although it converges faster, its total communication costs are slightly higher than {\tt{FedProto}}.}}

\begin{figure}[t]
\centering
\begin{minipage}[t]{0.44\linewidth}
\centering
\includegraphics[width=1.8in]{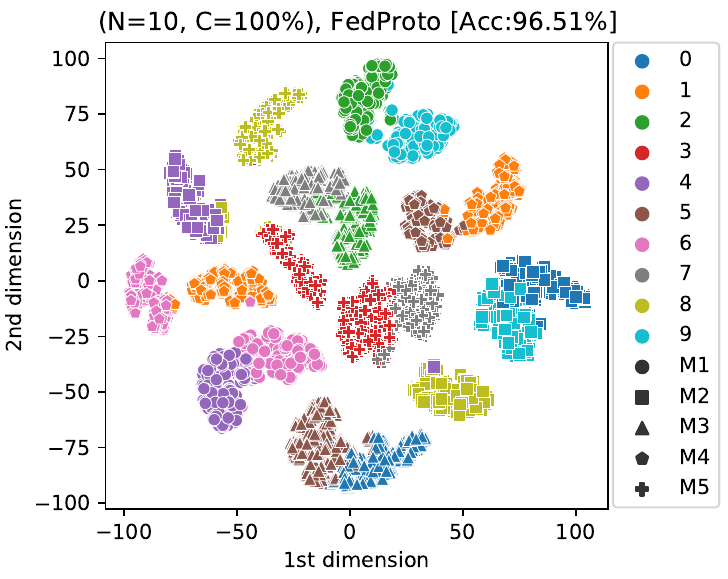}
\end{minipage}%
\begin{minipage}[t]{0.56\linewidth}
\centering
\includegraphics[width=1.8in]{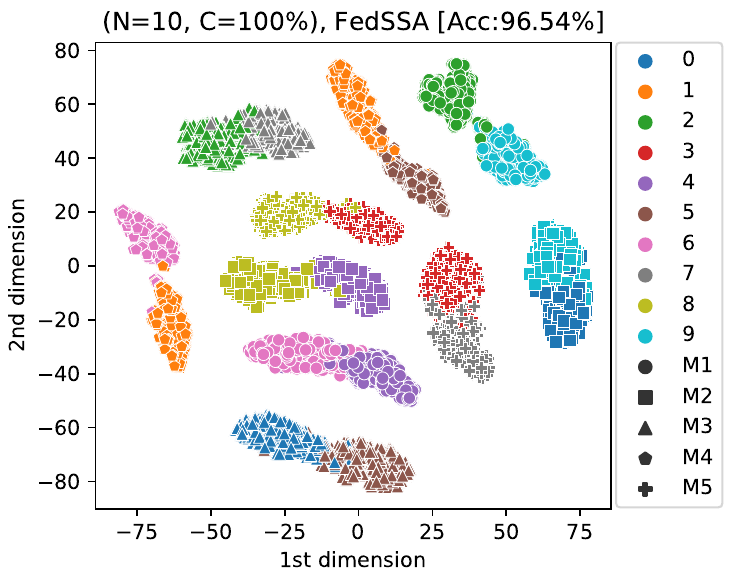}
\end{minipage}%


\vspace{-0.5em}
\caption{T-SNE representation visualization for {\tt{FedProto}} and \methodname{} on the CIFAR-10 (Non-IID: 2/10) dataset. {\ijcai{M1-M5 denote 5 heterogeneous models \{CNN-1, $\ldots$, CNN-5\}.}}}
\label{fig:compare-TSNE1}
\vspace{-1em}
\end{figure}

\textbf{Personalization Analysis}.
We extract every sample representation from each FL client under \methodname{} and {\tt{FedProto}}, respectively. Then, we leverage the T-SNE \cite{TSNE-JMLR} tool to reduce the dimensionality of the extracted representations from $500$ to $2$, and visualize the results. Since CIFAR-100 includes 100 classes of samples, we focus on visualizing the results on CIFAR-10 (non-IID: 2/10) in Figure~\ref{fig:compare-TSNE1}.
It displays that most clusters in \methodname{} and {\tt{FedProto}} consist of representations from a client's two seen classes of samples, indicating that each client's local model has strong personalization capability. 
Generally, \methodname{} performs better than {\tt{FedProto}}. 


\subsection{Case Studies}


\subsubsection{Robustness to Non-IIDness {\ijcai{(Class)}}}
\begin{figure}[ht!]
\centering
\begin{minipage}[t]{0.5\linewidth}
\centering
\includegraphics[width=1.7in]{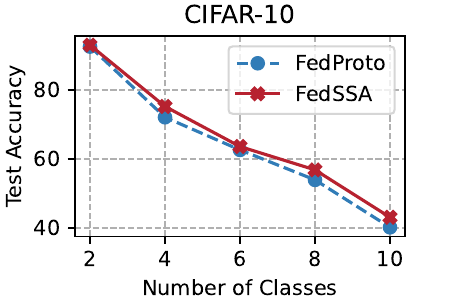}
\end{minipage}%
\begin{minipage}[t]{0.5\linewidth}
\centering
\includegraphics[width=1.7in]{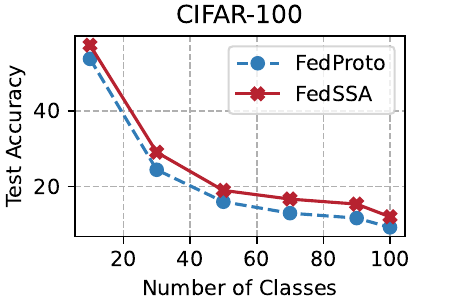}
\end{minipage}%
\vspace{-0.5em}
\caption{Robustness to Non-IIDness {\ijcai{(Class)}}.}
\label{fig:case-noniid}
\vspace{-0.5em}
\end{figure}

We study the robustness of \methodname{} and {\tt{FedProto}} to non-IIDness {\ijcai{controlled by the number of seen classes one client holds}} under (N=100, C=10\%) on CIFAR-10 and CIFAR-100. We vary the number of classes seen by each client as $\{2, 4, \ldots, 10\}$ on CIFAR-10 and $\{10, 30, \ldots, 100\}$ on CIFAR-100. Figure~\ref{fig:case-noniid} shows that \methodname{} consistently outperforms {\tt{FedProto}}, demonstrating its robustness to non-IIDness. As the non-IIDness decreases (the number of classes seen by each client rises), accuracy degrades as more IID local data enhances generalization and reduces personalization.

\begin{figure}[ht!]
\centering
\begin{minipage}[t]{0.5\linewidth}
\centering
\includegraphics[width=1.7in]{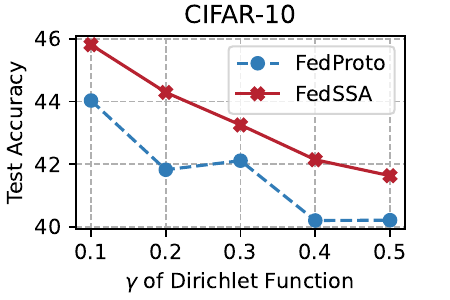}
\end{minipage}%
\begin{minipage}[t]{0.5\linewidth}
\centering
\includegraphics[width=1.7in]{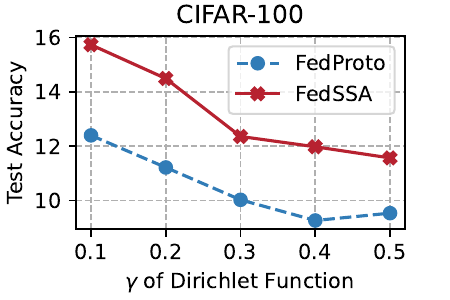}
\end{minipage}%
\vspace{-0.5em}
\caption{{\ijcai{Robustness to Non-IIDness (Dirichlet)}}.}
\label{fig:case-noniid-Dir}
\vspace{-0.5em}
\end{figure}

{\ijcai{
\subsubsection{Robustness to Non-IIDness (Dirichlet)}
We also test the robustness of \methodname{} and {\tt{FedProto}} to more complex non-IIDness controlled by a Dirichlet($\gamma$) function under setting (N=100, C=10\%) on CIFAR-10 and CIFAR-100. We vary $\gamma=\{0.1,\ldots,0.5\}$ on both datasets. Figure~\ref{fig:case-noniid-Dir} presents that \methodname{} also outperforms {\tt{FedProto}} across all non-IIDnesses. 
}}

\begin{figure}[ht!]
\centering
\begin{minipage}[t]{0.5\linewidth}
\centering
\includegraphics[width=1.7in]{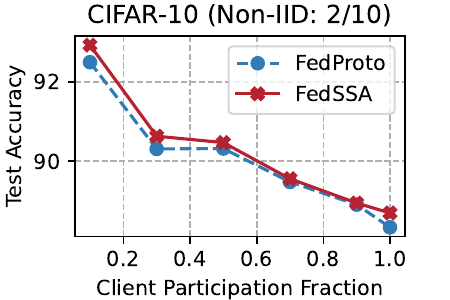}
\end{minipage}%
\begin{minipage}[t]{0.5\linewidth}
\centering
\includegraphics[width=1.7in]{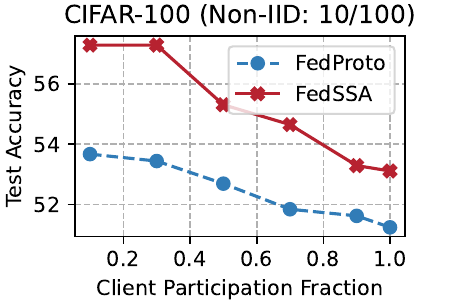}
\end{minipage}%
\vspace{-0.5em}
\caption{Robustness to client participation rates.}
\label{fig:case-frac}
\vspace{-0.5em}
\end{figure}

\subsubsection{Robustness to Client Participation Rates}
We also test the robustness of \methodname{} and {\tt{FedProto}} to client participation rates $C$ under (N=100, C=10\%) on CIFAR-10 (non-IID: 2/10) and CIFAR-100 (non-IID: 10/100). We vary the client participation rates as $C=\{0.1, 0.3, \ldots, 1\}$. Figure~\ref{fig:case-frac} shows that \methodname{} consistently outperforms {\tt{FedProto}}, especially on the more complicated CIFAR-100 dataset, demonstrating its robustness to changes in client participation rates. 

\section{Conclusions and Future Work} %
In this paper, we propose a novel personalized heterogeneous federated learning framework named \methodname{} to enhance the performance and efficiency of model-heterogeneous personalized federated learning. It consists of two core modules: 1) local-to-global knowledge transfer by semantic similarity-based header parameter aggregation and 2) global-to-local knowledge transfer by adaptive parameter stabilization-based header parameter fusion, both of them enhance the personalization of each client's heterogeneous local model and stabilize classification boundaries. Theoretical analysis shows that \methodname{} could converge over wall-to-wall time. Extensive experiments demonstrate that \methodname{} achieves the best classification accuracy while incurring the lowest communication and computational costs. 

In subsequent research, we will improve \methodname{} from two aspects: 1) designing a more effective aggregation rule for aggregating header parameters by class, and 2) extend {\methodname{}} to broader real-world applications.




\newpage

\section*{Acknowledgments}
This research is supported in part by 
the National Science Foundation of China under Grant 62272252 \& 62272253, the Key Research and Development Program of Guangdong under Grant 2021B0101310002, and the Fundamental Research Funds for the Central Universities;
National Key R\&D Program of China 2021YFF0900800 \& NSFC No.92367202;
the National Research Foundation Singapore and DSO National Laboratories under the AI Singapore Programme (AISG Award No: AISG2-RP-2020-019); the RIE 2020 Advanced Manufacturing and Engineering (AME) Programmatic Fund (No. A20G8b0102), Singapore;
the China Scholarship Council.


\bibliographystyle{named}
\bibliography{ijcai24}

\appendix
\onecolumn

\newpage
\section{Pseudo Code of \methodname{}}\label{app:alg}

\begin{algorithm}[H]

\caption{\methodname{}}
\label{alg:FedSSA}
\KwInput{
$N$, number of clients; $K$, number of selected clients in one round; $T$, total number of rounds; $\eta$, learning rate of local models; $\mu_0$, initial decay value; $T_{stable}$, number of rounds required for parameter stabilization. 
} 

Randomly initialize local personalized heterogeneous models $[f_0(\omega_0^0), f_1(\omega_1^0), \ldots, f_k(\omega_k^0), \ldots, f_{N-1}(\omega_{N-1}^0)] $ and the global homogeneous classification header $\mathcal{H}(\theta^0)$. \\ 
\For{each round t=0,1,..,T-1}{
    // \textbf{Server Side}: \\
    $\mathcal{K}^t$ $\gets$ Randomly sample $K$ clients from $N$ clients; \\
    Broadcast the global header $\theta^{t-1}$ to sampled $K$ clients; \\
    $\theta_k^{t,s}, s \in S_k, k \in \mathcal{K}^t \gets$ \textbf{ClientUpdate}($\theta^{t-1}$); \\
    \begin{tcolorbox}[colback=ylp_color2,
                  colframe=ylp_color1,
                  width=7.3cm,
                  height=2cm,
                  arc=1mm, auto outer arc,
                  boxrule=0.5pt,
                  left=0pt,right=0pt,top=0pt,bottom=0pt,
                 ]
    \For{$s \in S$}{
        /* Semantic Similarity-based Aggregation */ \\
        $\theta^{t,s}=\frac{1}{|\mathcal{K}_s^t|} \sum_{k \in \mathcal{K}_s^t} \theta_k^{t,s}$; \\
    }
    \end{tcolorbox}
    $\theta^t=\{\theta^{t,0},\theta^{t,1},\dots,\theta^{t,s},\dots,\theta^{t,S-1}\}$. \\
  \vspace{1em}
    // \textbf{ClientUpdate}: \\
    Receive the global header $\theta^{t-1}$ from the server; \\
    \For{$k \in \mathcal{K}^t$}{
    \begin{tcolorbox}[colback=ylp_color2,
                  colframe=ylp_color1,
                  width=7.8cm,
                  height=3.2cm,
                  arc=1mm, auto outer arc,
                  boxrule=0.5pt,
                  left=0pt,right=0pt,top=0pt,bottom=0pt,
                 ]
        Fuse local and global headers' seen-class parameters: \\
        \For{$s \in S_k$}{
              /* Adaptive Parameter Stabilization Strategy */ \\
            $\hat{\theta}_k^{t, s}=\theta^{t-1, s}+\mu^t \cdot \theta_k^{t-1, s}, s \in S_k$; \\
            $ \mu^t= \begin{cases}\mu_0 \cdot \cos (\frac{t}{2 T_{stable}} \pi), & t \leq T_{stable}; \\
            0, & t>T_{stable};\end{cases}$
         }
     \end{tcolorbox}
        Reconstruct local headers: 
        $\hat{\theta}_k^t=\{\hat{\theta}_k^{t, s}, \theta_k^{t-1, s^{\prime}}\}(s \in S_k, s^{\prime} \in (S \backslash S_k))$; \\
        Update local model: $\hat{\omega}_k^t = \varphi_k^{t-1} \circ \hat{\theta}_k^t$; \\
         Train local model by gradient descent: $\omega_k^t \gets \hat{\omega}_k^t-\eta \nabla(f_k(\boldsymbol{x}_i ; \hat{\omega}_k^t), y_i),(\boldsymbol{x}_i, y_i) \in D_k$; \\
        Upload seen-class parameters of trained local headers: $\theta_k^{t,s}, s \in S_k$. 
    }
}
\textbf{Return} personalized heterogeneous local models $[f_0(\omega_0^{T-1}), f_1(\omega_1^{T-1}), \ldots, f_k(\omega_k^{T-1}), \ldots, f_{N-1}(\omega_{N-1}^{T-1})]$.  
\end{algorithm}

\newpage
\section{Proof for Lemma~\ref{lemma:AfterAggregation}}\label{sec:proof-Lemma2}
\begin{proof}
\begin{equation}\label{eq:15}
\footnotesize
\begin{aligned}
\mathcal{L}_{(t+1) E+0}& =\mathcal{L}_{(t+1) E}+\mathcal{L}_{(t+1) E+0}-\mathcal{L}_{(t+1) E} \\
& \stackrel{(a)}{=} \mathcal{L}_{(t+1) E}+\mathcal{L}((\varphi_k^{t+1}, \hat{\theta}_k^{t+2}) ; \boldsymbol{x}, y)-\mathcal{L}((\varphi_k^{t+1}, \theta_k^{t+1}) ; \boldsymbol{x}, y) \\
& \stackrel{(b)}{\leq} \mathcal{L}_{(t+1) E}+\langle\nabla \mathcal{L} ((\varphi_k^{t+1}, \theta_k^{t+1})),((\varphi_k^{t+1}, \hat{\theta}_k^{t+2}) -(\varphi_k^{t+1}, \theta_k^{t+1}))\rangle +\frac{L_1}{2}\|(\varphi_k^{t+1}, \hat{\theta}_k^{t+2})-(\varphi_k^{t+1}, \theta_k^{t+1})\|_2^2 \\
& \stackrel{(c)}{\leq} \mathcal{L}_{(t+1) E}+\frac{L_1}{2}\|(\varphi_k^{t+1}, \hat{\theta}_k^{t+2})-(\varphi_k^{t+1}, \theta_k^{t+1})\|_2^2 \\
& \stackrel{(d)}{\leq} \mathcal{L}_{(t+1) E}+\frac{L_1}{2}\|\hat{\theta}_k^{t+2}-\theta_k^{t+1}\|_2^2 \\
& \stackrel{(e)}{=} \mathcal{L}_{(t+1) E}+\frac{L_1}{2}\|\hat{\theta}_k^{t+2, s}-\theta_k^{t+1, s}\|_{2, s \in S_k}^2 \\
& \stackrel{(f)}{=} \mathcal{L}_{(t+1) E}+\frac{L_1}{2}\|\theta^{t+1, s}+\mu^{t+2} \theta_k^{t+1, s}-\theta_k^{t+1, s}\|_{2, s \in S_k}^2 \\
& =\mathcal{L}_{(t+1) E}+\frac{L_1}{2}\|\theta^{t+1, s}+(\mu^{t+2}-1) \theta_k^{t+1, s}\|_{2, s \in S_k}^2{ }^{} \\
& \stackrel{(g)}{=} \mathcal{L}_{(t+1) E}+\frac{L_1}{2}\|\mathbb{E}_{k^{\prime} \in \mathcal{K}_s}[\theta_{k^{\prime}}^{t, s}-\eta \nabla \mathcal{L}(\theta_{k^{\prime}}^{t, s})] +(\mu^{t+2}-1)(\theta_k^{t, s}-\eta \nabla \mathcal{L}(\theta_k^{t, s}))\|_{2, s \in S_k}^2 \\
& =\mathcal{L}_{(t+1) E}+\frac{L_1}{2} \|(\mathbb{E}_{k^{\prime} \in \mathcal{K}_s}[\theta_{k^{\prime}}^{t, s}]+(\mu^{t+2}-1) \theta_k^{t, s}) -\eta(\mathbb{E}_{k^{\prime} \in \mathcal{K}_s}[\nabla \mathcal{L}(\theta_{k^{\prime}}^{t, s})]+(\mu^{t+2}-1) \nabla \mathcal{L}(\theta_k^{t, s})\|_{2, s \in S_k}^2 \\
& \stackrel{(h)}{\leq} \mathcal{L}_{(t+1) E}-\frac{\eta L_1}{2}\|\mathbb{E}_{k^{\prime} \in \mathcal{K}_s}[\nabla \mathcal{L}(\theta_{k^{\prime}}^{t, s})] +(\mu^{t+2}-1) \nabla \mathcal{L}(\theta_k^{t, s})\|_{2, s \in S_k}^2. 
\end{aligned}
\end{equation}

Take the expectation of $\mathcal{B}$ on both sides of Eq. \eqref{eq:15}, then:
\begin{equation}\label{eq:16}
\begin{aligned}
\mathbb{E}[\mathcal{L}_{(t+1) E+0}] & \leq \mathbb{E}[\mathcal{L}_{(t+1) E}] -\frac{\eta L_1}{2}\|\mathbb{E}_{k^{\prime} \in \mathcal{K}_s}[\nabla \mathcal{L}(\theta_{k^{\prime}}^{t, s})]  +(\mu^{t+2}-1) \nabla \mathcal{L}(\theta_k^{t, s})\|_{2, s \in S_k}^2 \\
& \stackrel{(i)}{\leq} \mathbb{E}[\mathcal{L}_{(t+1) E}]-\frac{\eta L_1 \delta^2}{2}. 
\end{aligned}
\end{equation}

In Eq.~\eqref{eq:15}, $(a)$: $\mathcal{L}_{(t+1) E+0}=\mathcal{L}((\varphi_k^{t+1}, \hat{\theta}_k^{t+2}); \boldsymbol{x}, y)$, i.e., at the start of the $(t+2)$-th round, the $k$-th client's local model consists of the local feature extractor $\varphi_k^{t+1}$ after local training in the $(t+1)$-th round, and the reconstructed local classification header $\hat{\theta}_k^{t+2}$ fused by the $(t+1)$-th round's local header $\theta_k^{t+1}$ and the $(t+1)$-th round's aggregated global header $\theta^{t+1}$ through the parameter stabilization strategy. $\mathcal{L}_{(t+1) E}=\mathcal{L}((\varphi_k^{t+1}, \theta_k^{t+1}) ; \boldsymbol{x}, y)$, i.e., in the $E$-th (last) local iteration of the $(t+1)$-th round, the $k$-th client's local model consists of the feature extractor $\varphi_k^{t+1}$ and the local classification header $\theta_k^{t+1}$. 
$(b)$ follows Assumption~\ref{assump:Lipschitz}. 
$(c)$: the inequality still holds when the second term is removed from the right-hand side. 
$(d)$: both $(\varphi_k^{t+1}, \hat{\theta}_k^{t+2})$ and $(\varphi_k^{t+1}, \theta_k^{t+1})$ have the same $\varphi_k^{t+1}$, the inequality still holds after it is removed.
$(e)$: the fused local classification header $\hat{\theta}_k^{t+2}$ consists of the mixed seen-class parameters $\hat{\theta}_k^{t+2,s}$ ($s\in S_k$) and unseen-class parameters from the historical local header $\theta_k^{t+1,s^{\prime}}$ ($s^{\prime} \in (S \backslash S_k)$), i.e., $\hat{\theta}_k^{t+2}=\{(\hat{\theta}_k^{t+2,s}|s\in S_k), (\theta_k^{t+1,s^{\prime}}|s^{\prime} \in (S \backslash S_k))\}$, and the historical local header $\theta_k^{t+1}=\{(\theta_k^{t+1,s}|s\in S_k),(\theta_k^{t+1,s^{\prime}}|s^{\prime} \in (S \backslash S_k))\}$, both of them involve $(\theta_k^{t+1,s^{\prime}}|s^{\prime} \in (S \backslash S_k))$, so $\|\hat{\theta}_k^{t+2}-\theta_k^{t+1}\|_2^2=\|\hat{\theta}_k^{t+2, s}-\theta_k^{t+1, s}\|_{2, s \in S_k}^2$.
$(f)$: as the parameter stabilization strategy defined in Eq.~(\ref{eq:para-stable}), $\hat{\theta}_k^{t+2, s}=\theta^{t+1, s}+\mu^{t+2} \theta_k^{t+1, s}, s \in S_k$.
$(g)$: the global header $\theta^{t+1, s}=\mathbb{E}_{k^{\prime} \in \mathcal{K}_s}[\theta_{k^{\prime}}^{t+1, s}]=\mathbb{E}_{k^{\prime} \in \mathcal{K}_s}[\theta_{k^{\prime}}^{t, s}-\eta \nabla \mathcal{L}(\theta_{k^{\prime}}^{t, s})]$ by  gradient descent. Similarly, $\theta_k^{t+1, s}=\theta_k^{t, s}-\eta \nabla \mathcal{L}(\theta_k^{t, s})$, $\eta$ is the learning rate of training local models.
$(h)$: since $\mu^{t+2}\in[0,1]$ and $\|\mathbb{E}_{k^{\prime} \in \mathcal{K}_s}[\theta_{k^{\prime}}^{t, s}]-\theta_k^{t, s}\|_{2, s \in S_k}^2 \leq \varepsilon^2$ defined in Assumption~\ref{assump:header}, $\|\mathbb{E}_{k^{\prime} \in \mathcal{K}_s}[\theta_{k^{\prime}}^{t, s}]+(\mu^{t+2}-1) \theta_k^{t, s}\|_{2, s \in S_k}^2\leq \varepsilon^2$, the inequality still holds after removing it from the right side. 
In Eq.~(\ref{eq:16}), 
{\ijcai{
$(i)$: for $\|\mathbb{E}_{k^{\prime} \in \mathcal{K}_s}[\nabla \mathcal{L}(\theta_{k^{\prime}}^{t, s})]+(\mu^{t+2}-1) \nabla \mathcal{L}(\theta_k^{t, s})\|_{2, s \in S_k}^2$, we use $X$ to denote $\nabla \mathcal{L}(\theta_{k^{\prime}}^{t, s})$, $X'$ to denote $\nabla \mathcal{L}(\theta_k^{t, s})$, and $a$ to denote $(1-\mu^{t+2})$, then we have $\|\mathbb{E}[X]-a X'\|_{2}^2$ which can be regarded a quadratic function $f(a)$ for $a$, \emph{i.e.}, $f(a)=(E[X]-aX')^2$.
Since the coefficient $X'^2$ of $a^2$ is larger than 0, $f(a)$ has a minimum value. Since $\mu \in (0,1)$, $a \in (0,1)$, the maximum value of $f(a)$ is determined $a=0$ or $a =1$. When $a =1$, $f(a)=(E[X]-X')^2 \leq \delta_1^2$, as assumed in Assumption~\ref{assump:header} ($\|\mathbb{E}_{k^{\prime} \in \mathcal{K}_s}[\nabla \mathcal{L}(\theta_{k^{\prime}}^{t, s})]-\nabla \mathcal{L}(\theta_k^{t, s})\|_{2, s \in S_k}^2 \leq \delta_1^2$).
When $a=0$, $f(a)=(E[X])^2$, since we clip gradient $X$ within a fixed range during model training, $(E[X])^2$ has boundary which can be considered as $\delta_2^2$, then $f(a)=(E[X])^2 \leq \delta_2^2$. It's not sure which is larger, $f(1)$ or $f(0)$, so we derive $f(a)=(E[X]-aX')^2 \leq \delta_1^2 + \delta_2^2 = \delta^2$. That is, $\|\mathbb{E}_{k^{\prime} \in \mathcal{K}_s}[\nabla \mathcal{L}(\theta_{k^{\prime}}^{t, s})]+(\mu^{t+2}-1) \nabla \mathcal{L}(\theta_k^{t, s})\|_{2, s \in S_k}^2 \leq \delta^2$.
}}
\end{proof}

\section{Proof for Theorem~\ref{theorem:one-round}}\label{sec:proof-Theorem1}
\begin{proof} 
By substituting Lemma~\ref{lemma:LocalTraining} into the second term on the right hand side of Lemma~\ref{lemma:AfterAggregation}, we have:
\begin{equation}
\begin{aligned}
\mathbb{E}[\mathcal{L}_{(t+1) E+0}] & \leqslant \mathcal{L}_{t E+0}-(\eta-\frac{L_1 \eta^2}{2}) \sum_{e=0}^E\|\mathcal{L}_{t E+e}\|_2^2 +\frac{L_1 E \eta^2}{2} \sigma^2 -\frac{\eta L_1 \delta^2}{2} \\
& \leqslant \mathcal{L}_{t E+0}-(\eta-\frac{L_1 \eta^2}{2}) \sum_{e=0}^E\|\mathcal{L}_{t E+e}\|_2^2 +\frac{\eta L_1(E \eta \sigma^2-\delta^2)}{2}
\end{aligned}
\end{equation}
\end{proof}

\section{Proof for Theorem~\ref{theorem:non-convex}}\label{sec:proof-Theorem2}
\begin{proof} 
Theorem~\ref{theorem:one-round} can be re-expressed as:
\begin{equation}\label{eq:18}
\begin{aligned}
\sum_{e=0}^E\|\mathcal{L}_{t E+e}\|_2^2 & \leqslant \frac{\mathcal{L}_{t E+0}-\mathbb{E}[\mathcal{L}_{(t+1) E+0}]}{\eta-\frac{L_1 \eta^2}{2}} + \frac{\frac{\eta L_1(E \eta \sigma^2-\delta^2)}{2}}{\eta-\frac{L_1 \eta^2}{2}}.
\end{aligned}
\end{equation}

Taking expectations of model $\omega$ on both sides of Eq. \eqref{eq:18}, then we can get:
\begin{equation}\label{eq:19}
\begin{aligned}
\sum_{e=0}^E \mathbb{E}[\|\mathcal{L}_{t E+e}\|_2^2] & \leqslant \frac{\mathbb{E}[\mathcal{L}_{t E+0}]-\mathbb{E}[\mathcal{L}_{(t+1) E+0}]}{\eta-\frac{L_1 \eta^2}{2}}  +\frac{\frac{\eta L_1(E \eta \sigma^2-\delta^2)}{2}}{\eta-\frac{L_1 \eta^2}{2}}
\end{aligned}
\end{equation}

Summing both sides over $T$ rounds ($t \in [0,T-1]$) yields:
\begin{equation}
\small
\begin{aligned}
\frac{1}{T} \sum_{t=0}^{T-1} \sum_{e=0}^E \mathbb{E}[\|\mathcal{L}_{t E+e}\|_2^2] & \leqslant \frac{\frac{1}{T} \sum_{t=0}^{T-1}(\mathbb{E}[\mathcal{L}_{t E+0}]-\mathbb{E}[\mathcal{L}_{(t+1) E+0}])}{\eta-\frac{L_1 \eta^2}{2}}  + \frac{\frac{\eta L_1(E \eta \sigma^2-\delta^2)}{2}}{\eta-\frac{L_1 \eta^2}{2}}.
\end{aligned}
\end{equation}

Since $\sum_{t=0}^{T-1}(\mathbb{E}[\mathcal{L}_{t E+0}]-\mathbb{E}[\mathcal{L}_{(t+1) E+0}]) \leqslant \mathcal{L}_{t=0}-\mathcal{L}^*$, then we can further derive:
\begin{equation} \label{eq:T-rounds}
\small
\begin{aligned}
 \frac{1}{T} \sum_{t=0}^{T-1} \sum_{e=0}^E \mathbb{E}[\|\mathcal{L}_{t E+e}\|_2^2] & \leqslant \frac{\frac{1}{T}(\mathcal{L}_{t=0}-\mathcal{L}^*)+\frac{\eta L_1(E \eta \sigma^2-\delta^2)}{2}}{\eta-\frac{L_1 \eta^2}{2}} \\
& =\frac{2(\mathcal{L}_{t=0}-\mathcal{L}^*)+\eta L_1 T(E \eta \sigma^2-\delta^2)}{T(2 \eta-L_1 \eta^2)}  \\
& =\frac{2(\mathcal{L}_{t=0}-\mathcal{L}^*)}{T \eta(2-L_1 \eta)}+\frac{L_1(E \eta \sigma^2-\delta^2)}{2-L_1 \eta}.
\end{aligned}
\end{equation}

If the local model can converge, the above equation satisfies the following condition:
\begin{equation}
\frac{2(\mathcal{L}_{t=0}-\mathcal{L}^*)}{\operatorname{T\eta }(2-L_1 \eta)}+\frac{L_1(E \eta \sigma^2-\delta^2)}{2-L_1 \eta} \leqslant \epsilon.
\end{equation}
Then, we can obtain:
\begin{equation}
T \geqslant \frac{2(\mathcal{L}_{t=0}-\mathcal{L}^*)}{\eta \epsilon(2-L_1 \eta)-\eta L_1(E \eta \sigma^2-\delta^2)}.
\end{equation}

Since $T>0, \mathcal{L}_{t=0}-\mathcal{L}^*>0$, we can further derive:
\begin{equation}
\eta \epsilon(2-L_1 \eta)-\eta L_1(E \eta \sigma^2-\delta^2) > 0,    
\end{equation}
i.e.,
\begin{equation}\label{eq:24}
\eta < \frac{2 \epsilon+L_1 \delta^2}{L_1(\epsilon+E \sigma^2)}.
\end{equation}

The right-hand side of Eq. \eqref{eq:24} are all constants. Thus, the learning rate $\eta$ is upper bounded. When $\eta$ satisfies the above condition, the second term of the right-hand side of Eq.~\eqref{eq:T-rounds} is a constant. It can be observed from the first term of Eq.~\eqref{eq:T-rounds} the non-convex convergence rate satisfies $\epsilon \sim \mathcal{O}(\frac{1}{T})$.
\end{proof}

\newpage
\section{Detailed Settings and Results in Experiments}\label{app:extra-exp}

\begin{table}[h!]
\centering
\caption{Structures of $5$ heterogeneous CNN models with $5 \times 5$ kernel size and $16$ or $32$ filters in convolutional layers.}
\resizebox{0.5\linewidth}{!}{%
\begin{tabular}{|l|c|c|c|c|c|}
\hline
Layer Name         & CNN-1    & CNN-2   & CNN-3   & CNN-4   & CNN-5   \\ \hline
Conv1              & 5$\times$5, 16   & 5$\times$5, 16  & 5$\times$5, 16  & 5$\times$5, 16  & 5$\times$5, 16  \\
Maxpool1              & 2$\times$2   & 2$\times$2  & 2$\times$2  & 2$\times$2  & 2$\times$2  \\
Conv2              & 5$\times$5, 32   & 5$\times$5, 16  & 5$\times$5, 32  & 5$\times$5, 32  & 5$\times$5, 32  \\
Maxpool2              & 2$\times$2   & 2$\times$2  & 2$\times$2  & 2$\times$2  & 2$\times$2  \\
FC1                & 2000     & 2000    & 1000    & 800     & 500     \\
FC2                & 500      & 500     & 500     & 500     & 500     \\
FC3                & 10/100   & 10/100  & 10/100  & 10/100  & 10/100  \\ \hline
model size & 10.00 MB & 6.92 MB & 5.04 MB & 3.81 MB & 2.55 MB \\ \hline
\end{tabular}%
}
\label{tab:model-structures}
\end{table}

\begin{figure}[h!]
\centering
\begin{minipage}[t]{0.25\textwidth}
\centering
\includegraphics[width=1.8in]{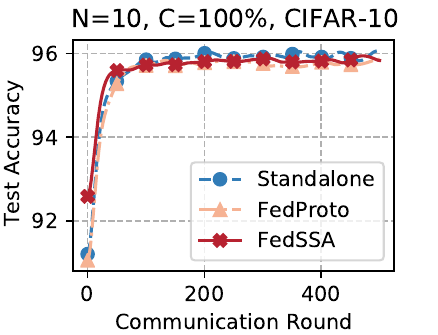}
\end{minipage}%
\begin{minipage}[t]{0.25\textwidth}
\centering
\includegraphics[width=1.8in]{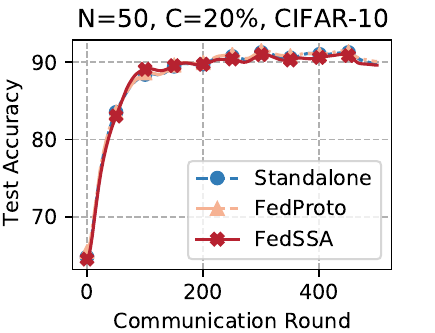}
\end{minipage}%
\begin{minipage}[t]{0.25\textwidth}
\centering
\includegraphics[width=1.8in]{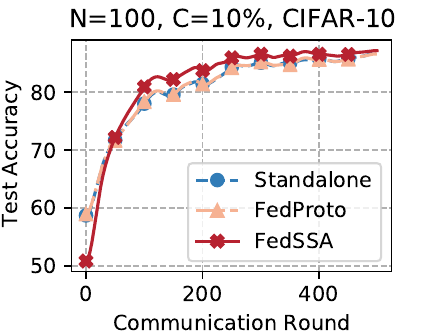}
\end{minipage}%

\begin{minipage}[t]{0.25\textwidth}
\centering
\includegraphics[width=1.8in]{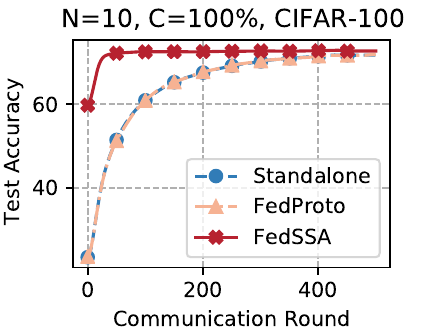}
\end{minipage}%
\begin{minipage}[t]{0.25\textwidth}
\centering
\includegraphics[width=1.8in]{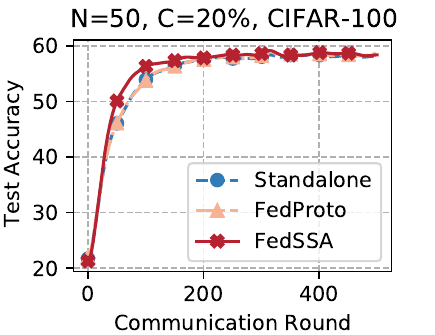}
\end{minipage}%
\begin{minipage}[t]{0.25\textwidth}
\centering
\includegraphics[width=1.8in]{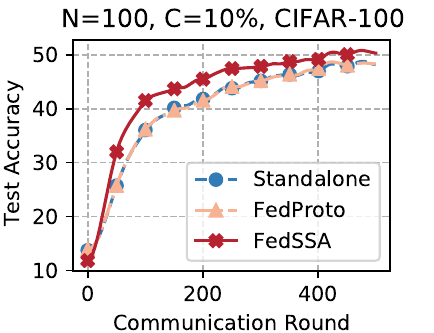}
\end{minipage}%
\caption{Mean test accuracy after smooth processing vs. communication rounds in \textit{model-heterogeneous} scenarios. }
\label{fig:compare-hetero-converge}
\end{figure}



\newpage
\section{Ablation Studies}

\methodname{} consists of two core modules: 1) the semantic similarity-based header parameter aggregation module (denoted as `M1') for local-to-global knowledge transfer, and 2) the adaptive parameter stabilization-based header parameter fusion module (denoted as `M2') for global-to-local knowledge transfer.
{\ijcai{We conduct ablation experiments for `M1' and `M2', respectively.}}

\begin{table}[h]
\centering
\caption{{\ijcai{Ablation study results for `M1'.}}}
\label{tab:ablation-M1}
\resizebox{0.7\linewidth}{!}{%
\begin{tabular}{|ccc|cc|cc|cc|}
\hline
     &                            &                            & \multicolumn{2}{c|}{N=10, C=100\%} & \multicolumn{2}{c|}{N=50, C=20\%} & \multicolumn{2}{c|}{N=100, C=10\%} \\ \cline{4-9} 
Case & M1                    & M2                    & CIFAR10         & CIFAR100       & CIFAR10         & CIFAR100      & CIFAR10         & CIFAR100       \\ \hline
A    & \ding{55} & \ding{55} & 96.30            & 72.20           & 94.83            & 60.95           & 91.27            & 45.83           \\
B    & \ding{55} & \ding{51} & \cellcolor[HTML]{D3D3D3}96.42            & \cellcolor[HTML]{D3D3D3}72.69           & \cellcolor[HTML]{D3D3D3}95.10            & \cellcolor[HTML]{D3D3D3}62.53          & \cellcolor[HTML]{D3D3D3}91.85            & \cellcolor[HTML]{D3D3D3}50.71           \\
C    & \ding{51} & \ding{51} & \cellcolor[HTML]{9B9B9B}\textbf{96.54}   & \cellcolor[HTML]{9B9B9B}\textbf{73.39}  & \cellcolor[HTML]{9B9B9B}\textbf{95.83}   & \cellcolor[HTML]{9B9B9B}\textbf{64.20}  & \cellcolor[HTML]{9B9B9B}\textbf{92.92}   & \cellcolor[HTML]{9B9B9B}\textbf{57.29}  \\ \hline
\end{tabular}%
}
\end{table}

{\ijcai{\textbf{Ablation for `M1'.} We consider three cases for `M1': A) {\tt{LG-FedAvg}} aggregates whole local headers to generate the global header; B) {\tt{LG-FedAvg}} further apply adaptive parameter stabilization (`M2') on clients; C) \methodname{} with both `M1' and `M2'.
Table~\ref{tab:ablation-M1} presents that the adaptive parameter stabilization strategy further improves {\tt{LG-FedAvg}}, but it still performs worse than \methodname{}, indicating the necessity of combining `M1' and `M2'. 
}}

\begin{table}[h]
\centering
\caption{Ablation study results for `M2'.}
\label{tab:ablation-M2}
\resizebox{0.7\linewidth}{!}{%
\begin{tabular}{|ccc|cc|cc|cc|}
\hline
     &                            &                            & \multicolumn{2}{c|}{N=10, C=100\%} & \multicolumn{2}{c|}{N=50, C=20\%} & \multicolumn{2}{c|}{N=100, C=10\%} \\ \cline{4-9} 
Case & M1                    & M2                    & CIFAR10         & CIFAR100       & CIFAR10         & CIFAR100      & CIFAR10         & CIFAR100       \\ \hline
A    & \ding{51} & \ding{55} & \cellcolor[HTML]{D3D3D3}96.36            & \cellcolor[HTML]{D3D3D3}73.26           & \cellcolor[HTML]{D3D3D3}92.68            & \cellcolor[HTML]{D3D3D3}58.01          & \cellcolor[HTML]{D3D3D3}88.51            & \cellcolor[HTML]{D3D3D3}49.95           \\
B    & \ding{51} & \ding{55} & 93.01            & 71.54           & 82.42            & 52.14          & 80.41            & \cellcolor[HTML]{D3D3D3}49.95           \\
C    & \ding{51} & \ding{51} & \cellcolor[HTML]{9B9B9B}\textbf{96.54}   & \cellcolor[HTML]{9B9B9B}\textbf{73.39}  & \cellcolor[HTML]{9B9B9B}\textbf{95.83}   & \cellcolor[HTML]{9B9B9B}\textbf{64.20}  & \cellcolor[HTML]{9B9B9B}\textbf{92.92}   & \cellcolor[HTML]{9B9B9B}\textbf{57.29}  \\ \hline
\end{tabular}%
}
\end{table}

{\ijcai{\textbf{Ablation for `M2'.}}}
We {\ijcai{use `M1' and}} discuss three cases {\ijcai{for `M2'}}: A) similar to {\tt{LG-FedAvg}}, each client replaces its historical local header with the latest global header; B) each client only replaces the seen-class parameters of its historical local header with those of the latest global header; and C) each client follows \methodname{}. We evaluate these three cases under model-heterogeneous settings.

\begin{figure}[h]
\centering
\begin{minipage}[t]{0.5\linewidth}
\centering
\includegraphics[width=0.6\linewidth]{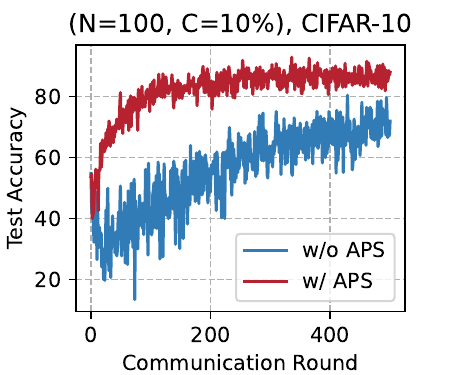}
\end{minipage}%
\begin{minipage}[t]{0.5\linewidth}
\centering
\includegraphics[width=0.6\linewidth]{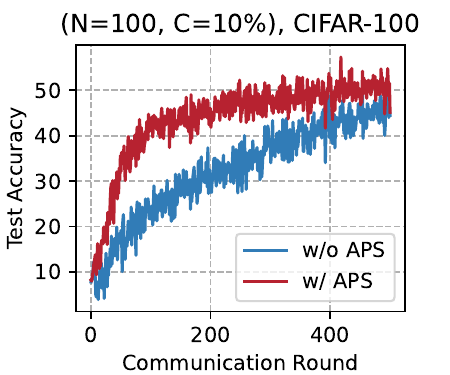}
\end{minipage}%
\caption{{\ijcai{Test accuracy varies as communication round for \methodname{} without or with adaptive parameter stabilization strategy.}}}
\label{fig:ablation-shaking}
\end{figure}

Table~\ref{tab:ablation-M2} shows that Case-C achieves the best performance, followed by Case-A and Case-B, demonstrating the effectiveness of the proposed header parameter fusion with the adaptive parameter stabilization strategy. The reason is that Case-C integrates the local and global knowledge of seen classes, but Case-B ignores seen-class local knowledge. Case-A not only omits seen-class local knowledge but also includes unseen-class global knowledge that may disturb the classification of seen classes. {\ijcai{Since Case-A exchanges the whole local or global header between the server and clients, it improves the generalization to all classes but compromises the personalization to local seen classes.}}
{\ijcai{Besides, Figure~\ref{fig:ablation-shaking} shows that \methodname{} with `M2' (Case-C) indeed can mitigate parameter shaking and speed convergence compared with not using `M2' (Case-B), especially on CIFAR-10.
}}

\end{document}